\documentclass{article}

\usepackage{ijcai16}

\usepackage{times}
\usepackage{epsfig}
\usepackage{graphicx}
\usepackage{amsmath}
\usepackage{amssymb}
\usepackage{mathsymb}
\usepackage{textcomp}
\usepackage[linesnumbered,algoruled,boxed,lined]{algorithm2e}
\usepackage{verbatim}
\usepackage{subfigure}
\usepackage{url}
\usepackage{graphicx}
\usepackage{multirow}
\usepackage[super]{nth}
\usepackage{balance}
\usepackage{colortbl}
\graphicspath{{./}{./Fig/}{./Figs/}{./Fig/eps/}{./Fig/pdf/}}

\usepackage{algorithm2e}

\let\savedalgorithm\algorithm
\let\savedendalgorithm\endalgorithm

\usepackage{algorithm}

\newtheorem{thm}{Theorem}      
\newtheorem{exam}{Example}

\newcommand{\etal}{\textit{et al.\ }}
\newcommand{\eg}{\emph{e.g.,\ }}
\newcommand{\ie}{\emph{i.e.,\ }}

\def\bw{{\boldsymbol w}}

\title{Structured Deep Hashing with Convolutional Neural Networks for Fast Person Re-identification}
\author{Lin Wu$^{\ddag}$, Yang Wang$^{\dag}$\\
 $^{\ddag}$ISSR, ITEE, The University of Queensland, Brisbane, QLD, 4072, Australia\\
 $^{\dag}$The University of New South Wales, Kensington, Sydney, Australia\\
 lin.wu@uq.edu.au \space\space wangy@cse.unsw.edu.au
}

\begin{document}

\maketitle

\begin{abstract}

Given a pedestrian image as a query, the purpose of person re-identification is to identify the correct match from a large collection of gallery images depicting the same person captured by disjoint camera views. The critical challenge is how to construct a robust yet discriminative feature representation to capture the compounded variations in pedestrian appearance. To this end, deep learning methods have been proposed to extract hierarchical features against extreme variability of appearance. However, existing  methods in this category generally neglect the efficiency in the matching stage whereas the searching speed of a re-identification system is crucial in real-world applications.
In this paper, we present a novel deep hashing framework with Convolutional Neural Networks (CNNs)
for fast person re-identification. Technically, we simultaneously learn both CNN features and hash functions/codes to get robust yet discriminative features and similarity-preserving hash codes. Thereby, person re-identification can be resolved by efficiently computing and ranking the Hamming distances between images. A structured loss function defined over positive pairs and hard negatives is proposed to formulate a novel optimization problem so that fast convergence and more stable optimized solution can be obtained. Extensive experiments on two benchmarks CUHK03 \cite{FPNN} and Market-1501 \cite{Market1501} show that the proposed deep architecture is efficacy over state-of-the-arts.

\end{abstract}

\section{Introduction}\label{sec:intro}

re-identification is task of matching persons observed from non-overlapping camera views based on visual appearance. It has gained considerable popularity in video surveillance, multimedia, and security system by its prospect of searching a person of interest from a large amount of video sequences \cite{Zero-shot-person,Person-ranking,Person-context}.
The major challenge arises from the variations in human appearances, poses, viewpoints and background cluster across camera views. Some examples are shown in Fig.\ref{fig:example}. Towards this end, many approaches \cite{Farenzena2010Person,Yang:color-person,MidLevelFilter,Pedagadi2013Local,LADF,paul2015ensemble} have been proposed by developing a combination of low-level features (including color histogram \cite{Gray2008Viewpoint}, spatial co-occurrence representation \cite{WangShape2007}, LBP \cite{Xiong2014Person} and color SIFT \cite{eSDC}) against variations (\eg poses and illumination) in pedestrian images.
However, these hand-crafted features are still not discriminative and reliable to such severe variations and misalignment across camera views.


Recently, deep learning methods \cite{FPNN,JointRe-id,DeepReID,Ding:DeepRank,PersonNet,DeepRanking,DomainDropout} have been proposed to address the problem of person re-identification by learning deeply discriminative Convolutional Neural Network (CNN) features in a \emph{feed-forward} and \emph{back-propagation} manner. It extracts hierarchical CNN features from pedestrian images; the subsequent metric-cost part compares the CNN features with a chosen metric encoded by specific loss functions, \eg contrastive (pair-wise) \cite{FPNN,JointRe-id,PersonNet} or triplet \cite{DeepReID,DeepRanking} loss functions. However, such typical deep learning methods are not efficient in real-time scenario, due to the less-efficiency of matching two pedestrian images by extracting and comparing hierarchical CNN features. In fact, the excellent recognition accuracy in neural network-based architectures comes at expense of high computational cost both at training and testing time. The main computational expense for these deep models comes from convolving filter maps with the entire input image, making their computational complexity at least linear in the number of pixels. And matching these CNN features to obtain similarity values is not fast enough to be applicable in real-world applications. In this paper, we aim to reduce the computational burden of person re-identification by developing a fast re-identification framework.

\subsection{Motivation}
To cope with ever-growing amounts of visual data, deep learning based hashing methods have been proposed to simultaneously learn similarity-preserved hashing functions and discriminative image representation via a deep architecture \cite{Lai:hash-deep,DeepSemanticHash,DeepRegularlizedHash}. Simply delving existing deep hashing approaches into a person re-identification system is not trivial due to the difficulty of generalizing these pre-trained models to match pedestrian images in disjoint views.
Fine-tuning is a plausible way to make pre-trained models suitable to re-identification, however, to suit their models, training images are commonly divided into mini-batches, where each mini-batch contains a set of \textit{randomly} sampled positive/negative pairs or triplets. Thus, a contrastive or triplet loss is computed from each mini-batch, and the networks try to minimize the loss function feed-forwardly and update the parameters through back-propagation by using Stochastic Gradient Decent (SGD) \cite{SGD}.

We remark that randomly sampled pairs/triplets carry little helpful information to SGD. For instance, many triplet units can easily satisfy the relative comparison constraint in a triplet loss function (Eq \eqref{eq:triplet_constraint}), resulting into a slow convergence rate in the training stage.
Worse still, mini-batches with random samples may fail to obtain a stable solution or collapsed into a local optimum if a contrastive/triplet loss function is optimized \cite{LiftEmbed}. To this end, a suitable loss function is highly demanded to work well with SGD over mini-batches.

In this paper, we propose a deep hashing based on CNNs to efficiently address the problem of person re-identification. To mitigate the undesirable effects caused by contrastive/triplet loss function, we propose a structured
loss function by \textit{actively} adding hard negative samples into mini-batches, leading to a structured deep hashing framework. The proposed structured loss can guide sub-gradient computing in SGD to have correct directions, and thus achieves a fast convergence in training.

\subsection{Our Approach}
One may easily generate a straightforward two-stage deep hashing strategy by firstly extracting CNN features from a pre-trained model \eg AlexNet \cite{Krizhevsky2012Imagenet}, followed by performing the learned hash functions (separate projection and quantization step) to convert such CNN features into binary codes. However, as demonstrated in section \ref{sec:exp},  such a strategy cannot obtain optimal binary codes. As such binary codes may not well characterize the supervised information from training data \ie intra-personal variation and inter-personal difference, due to the \emph{independence} of two stages. In fact, such two stages can boost each other to achieve much better performance, that is, the learned binary codes can guide the learning of useful CNN features, while CNN features can in turn help learn semantically similarity-preserving hash function/codes.

\begin{figure}[t]
    \centering
        \includegraphics[width=3.5in,height=2in]{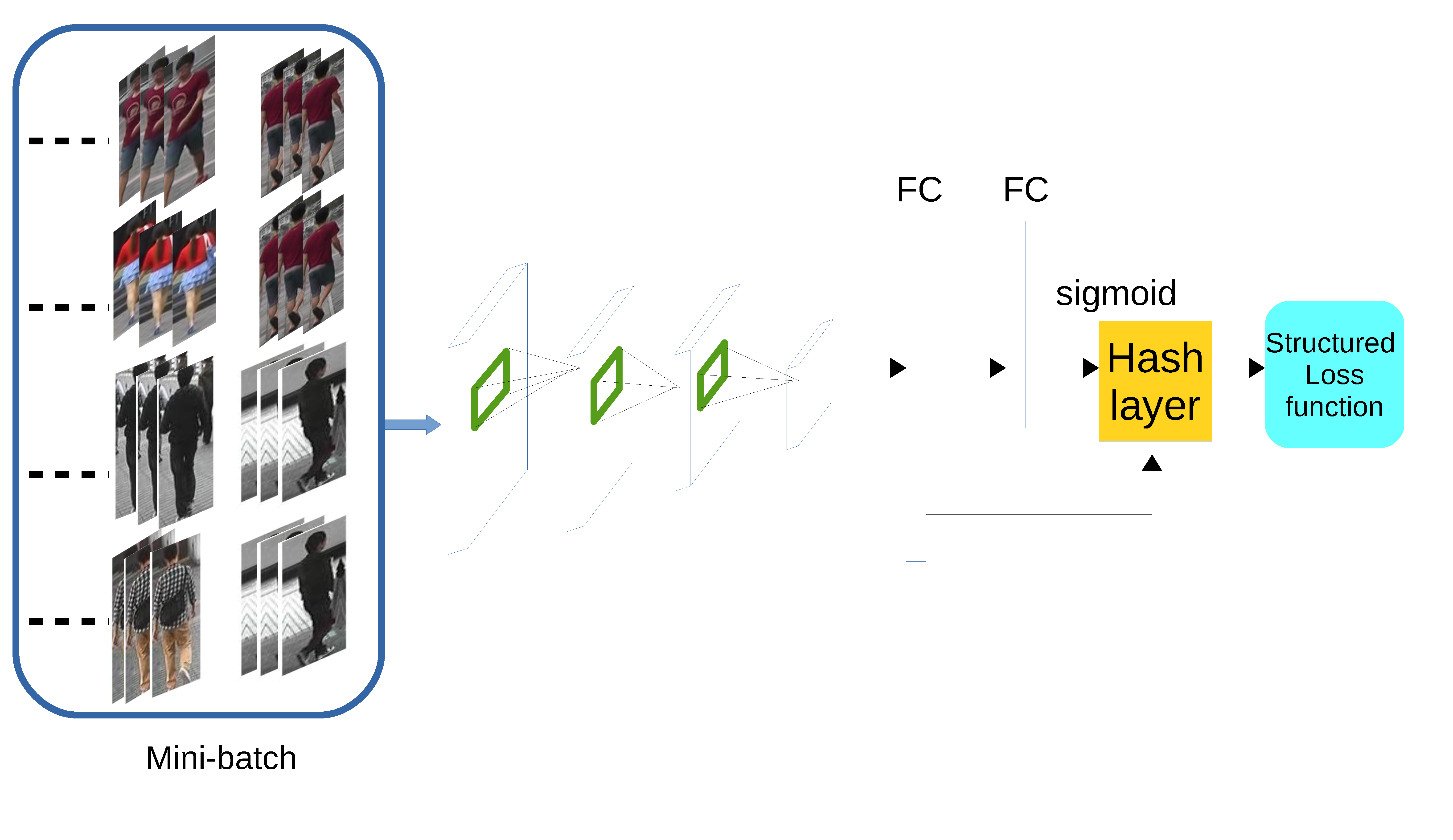}
    \caption{Overview of our deep hashing framework for person re-identification.  Our deep neural network takes a feed-forward, back-propagation strategy to learn features and hash codes simultaneously. During the feed-forward stage, the proposed network performs inference from  a mini-batch. The mini-batch is put through a stack of convolutional layers to generate nonlinear yet discriminative features, which are subsequently mapped to output feature vector by fully-connected layers (FC).  Meanwhile, a hash function layer is introduced at the top of FC layer to learn hash codes that are optimized by a structured loss function to preserve their similarities/dissimilarities. In back-propagation,  parameters are updated by computing their Stochastic Gradient Decent (SGD) w.r.t. the mini-batch.}\label{fig:framework}
\end{figure}

Motivated by this, we present a structured deep hashing architecture to \textit{jointly} learn feature representations and hash codes for person re-identification. The overall framework is illustrated in Fig.\ref{fig:framework}. In our architecture, mini-batches contain all positive pairs for a particular pedestrian, meanwhile each positive pair (has a query image and its correct match image from a different camera view) is augmented by actively selected hard negatives for its query and match image, respectively. Such mini-batches are taken into the inputs of deep network with a structured loss function optimized to learn CNN features and hash functions jointly.

The major contributions are summarized below:
\begin{itemize}
\item To the best of our knowledge, we are the first to solve person re-identification efficiently by presenting a structured deep hashing model. This makes our paper distinct from existing studies \cite{Zero-shot-person,Person-ranking,Person-context} where the matching efficiency is not addressed.
\item By simultaneously learning CNN features and hash functions/codes, we are able to get robust yet discriminative features against complex pedestrian appearance and boosted hash codes, so that every two deep hashing codes learned from the same identities are close to each other while those from different identities are kept away.
\item To combat the drawbacks of the contrastive/triplet loss, we propose a  structured loss function where mini-batches are augmented by considering hard negatives. Also, the proposed structured loss function that is imposed at the top layer of the network can achieve fast convergence and a stable optimized solution.
\end{itemize}

\section{Related Work}\label{sec:related}

In this section, we briefly review deep learning based on CNNs for person re-identification and several typical hashing methods, as they are closely related to our proposed technique.

In literature of person re-identification, many studies try to address this challenging problem by either seeking a robust feature representation \cite{Farenzena2010Person,MidLevelFilter,YangCIKM13,LinMM13,YangKAIS16,YangMM15,YangTIP15,TNNLS17,PAKDD14,YangIJCAI16,Gray2008Viewpoint,eSDC} or casting it as a metric learning problem where more discriminative distance metrics are learned to handle features extracted from person images across camera views \cite{LADF,KISSME,Pedagadi2013Local,Xiong2014Person,LOMOMetric,NullSpace-Reid}. The first aspect considers to find features that are robust to challenging factors while preserving identity information. The second stream generally tries to minimize the intra-class distance while maximize the inter-class distance. Also, person re-identification can be approached by a pipeline of image search where a Bag-of-words \cite{Market1501} model is constructed to represent each pedestrian image and visual matching refinement strategies can be applied to improve the matching precision. Readers are kindly referred to \cite{PersonReid-book} to have more reviews.

A notable improvement on person re-identification is achieved by using Convolutional Neural Networks (CNNs) \cite{FPNN,JointRe-id,DeepReID,Ding:DeepRank,PersonNet,Lin-PR2016,DeepRanking,Yang-TIP2017,DomainDropout}, which can jointly learn robust yet discriminative feature representation and its corresponding similarity value in an end-to-end fashion.
However, existing deep learning methods in person re-identification are facing a major challenge of efficiency, where computational time required to process an input image is very high due to the convolution operations with the entire input through deep nets.
Thus, from a pragmatical perspective, an advanced yet fast neural network-based architecture is highly demanded. This motivated us to develop an efficient deep learning model to alleviate the computational burden in person re-identification.

Hashing is an efficient technology in approximate nearest neighbor search with low storage cost of loading hash codes.
Learning-based hash methods can be roughly divided into two categories: unsupervised methods and supervised methods.
Unsupervised methods including Spectral Hashing \cite{Weiss:SH,Yang-SIGIR15} and Iterative Quantization \cite{Gong:cvpr2011} only use the training data to learn hash functions. Supervised methods try to leverage supervised information to learn compact binary codes. Some representative methods are Binary Reconstruction Embedding (BRE) \cite{Kulis:nips2009}, Minimal Loss Hashing (MLH) \cite{Norouzi:loss-hash}, and Supervised Hashing with Kernels (KSH) \cite{Liu:kernel-hash}.

Nonetheless, these hashing methods often cope with images represented by hand-crafted features (\eg SIFT \cite{eSDC}), which are extracted before projection and quantization steps. Moreover, they usually seek a linear projection which cannot capture the nonlinear relationship of pedestrian image samples\footnote{Pedestrian images typically undergo compounded variations in the form of human appearance, view angles, and human poses.}.  Even though some kernel-based hashing approaches \cite{Liu:kernel-hash,Lin-IVC2017} have been proposed, they are stricken with the efficiency issue.
To capture the non-linear relationship between data samples while keeping efficient, Liong \etal \cite{liong:deep}  present a Deep Hashing to learn multiple hierarchical nonlinear transformation which maps original images to compact binary code and thus supports large-scale image retrieval. A supervised version named Semantic Deep Hashing is also presented in \cite{liong:deep} where a discriminative item is introduced into the objective function. However, the above methods did not include a pre-training stage in their networks, which may make the generated hash codes less semantic. To keep the hash codes semantic, Xia \etal \cite{Xia:hash2014} proposed a deep hashing architecture based on CNNs, where the learning process is decomposed into a stage of learning approximate hash codes from supervised priors, which are used to guide a stage of simultaneously learning hash functions and image representations.

More recently, to generate the binary hash codes directly from raw images, deep CNNs are utilized to train the model in an end-to-end manner where discriminative features and hash functions are simultaneously optimized \cite{Lai:hash-deep,DeepSemanticHash,DeepRegularlizedHash}. However, in training stage, they commonly take mini-batches with randomly sampled triplets as inputs, which may lead to local optimum or unstable optimized solution.

By contrast, in this paper we deliver the first efforts in proposing a structured deep hashing model for person re-identification, which allows us to jointly learn deep feature representations and binary codes faithfully. The proposed structured loss function benefits us from achieving fast convergence and more stable optimized solutions, compared with pairwise/triplet ranking loss.
\begin{figure}[t]
    \centering
        \includegraphics[width=3.5in,height=1.2in]{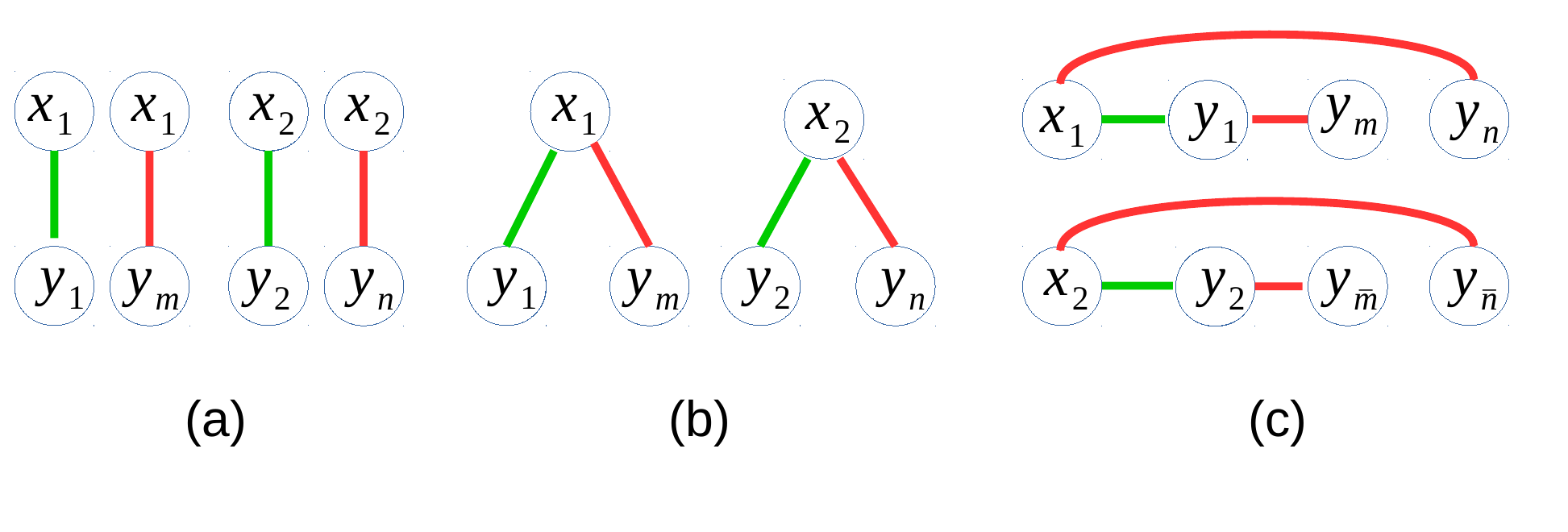}
    \caption{Illustrations on different loss functions. (a) Contrastive loss; (b) Triplet ranking loss; (c) Our structured loss. Here, $\bx$'s and $\by$'s indicate hash codes of pedestrian images captured by query and gallery camera view, respectively. For a specific pedestrian's hash codes $\bx_i$, its correct match's code is $\by_i$ from a different view.  Green edges and red edges represent similar and dissimilar examples, respectively. Our method explicitly adds hard negatives (\eg $y_m$, $y_n$) for all positive pairs (\eg ($x_1, y_1$) and ($x_2$, $y_2$) ) into mini-batches.}\label{fig:loss}
\end{figure}

\section{Structured Deep Hashing for Person Re-identification}\label{sec:method}
Our major contribution is to jointly learn feature representation from raw person images and their mappings to hash codes by presenting an improved deep neural network. The proposed network takes a mini-batch as its input which contains images in a form of positive/negative pairs. The architecture consists of three components: 1)  a stack of convolution layers followed by max pooling to learn non-linear feature mappings from raw pedestrian images; 2) a hash layer connected to the first and the second fully connected layers; 3) a structured loss function is designed to optimize the whole mini-batch. The architecture overview is illustrated in Fig.\ref{fig:framework}.

\subsection{Learning Deep Hashing Functions}
Assuming $\mathcal{I}$ to be the original image space, a hash function $f: \mathcal{I} \rightarrow \{0,1\}^r$ is treated as a mapping that projects an input image $\mathcal{I}$ into a $r$-bit binary code $f(\mathcal{I})$ while preserving the similarities of person images across camera views.

Learning based hashing methods aim to seek a set of hash functions to map and quantize each sample into a compact binary code vector. Assuming we have $r$ hash functions to be learned, which map an image $\mathcal{I}$ into a $r$-bit binary code vector $f(\mathcal{I})=[f_1(\mathcal{I}), f_2(\mathcal{I}), \ldots, f_r(\mathcal{I})]$.
Although many learning-based hashing methods have been proposed \cite{Gong:angular,Gong:cvpr2011,He:kmeans-hash,Norouzi:loss-hash,Kulis:nips2009}, most of them essentially learn a single linear projection matrix, which can not well capture the nonlinear relationship of samples. Admittedly, some kernel-based hashing methods are available \cite{Liu:kernel-hash,He:kdd2010}, they instead suffer from the efficiency issue because kernel-based methods cannot have explicit nonlinear mapping.

In this work, we propose to learn deep hash functions with CNNs to jointly learn feature representation from raw pixels of pedestrian images and their mappings to hash codes. In this way, feature representations for person images can be learned more optimally compatible with the coding process, thus producing optimal hash codes.

During training, the input to our network is a mini-batch containing pairs of fixed-size 160$\times$60 RGB images. The images are passed through four convolutional layers, where we use filters with a very small receptive filed: 3$\times$3. The convolution stride is fixed to 1 pixel. Spatial pooling is carried out by three max-pooling layers. Max-pooling is performed over a 2$\times$2 pixel window, with stride 2. After a stack of convolution layers, we have two fully-connected layers where the first one has 4096 dimension and the second is 512-dim, which are then fed into the hash layer to generate a compact binary code. We show details of layers in CNNs in Table \ref{tab:layer}.

\begin{table}[t]
  \centering
  \caption{Layer parameters of convolutional neural networks. The output dimension is given by height$\times$width$\times$width. FS: filter size for convolutions. Layer types: C: convolution, MP: max-pooling, FC: fully-connected.  All convolution and FC layers use hyperbolic tangent as activation function.}  \label{tab:layer}
  {
  \begin{tabular}{c|c|c|c|c}
  \hline
\hline
    Name  & Type  &  Output Dim & FS  & Sride \\
  \hline\hline
   Conv0  & C & $157\times57\times32$  & 3$\times3$ & 1 \\
   Pool0  & MP &  $79\times 29\times 32$ & 2$\times2$ & 2 \\
   Conv1  & C & $76\times 26\times 32$ & 3$\times$3 & 1 \\
   Pool1  & MP & $38\times 13\times 32$ & 2$\times$2 & 2 \\
   Conv2  & C &  $35\times 10\times 32$ &  3$\times$3  & 1 \\
   Pool2  & C & $18\times 5\times 32$ & 3$\times$3 &  2\\
   Conv3  & C & $15\times 2\times 32$ & 3$\times$3 &  1\\
   Pool4  & MP & $15\times 2\times 32$ & 1$\times$1 &  1\\
\hline
   FC1   & FC  & -  & 4096 &  - \\
   FC2   & FC  & - &  512 & -  \\
  \hline
  \end{tabular}
  }
\end{table}

Inspired by \cite{Sun:deep-face}, we add a bypass connection between the first fully connected layer and the hash layer to reduce the possible information loss. Another reason is features from the second fully connected layer is very semantic and invariant, which is unable to capture the subtle difference between person images. Thus, we rewrite the deep hash function as:
\begin{equation}
f(\mathcal{I},\bw_i)=sigmoid\left(\bw_i^T [g_1(\mathcal{I}); g_2(\mathcal{I})]\right),
\end{equation}
where $sigmoid(t)=1/(1+\exp^{(-\bw^T t)})$, $\bw_i$ denotes the weights in the $i$-th hash function, $g_1(\cdot)$ and $g_2(\cdot)$ represent feature vectors from the outputs of the two fully connected layers, respectively. Then, we have $f(\mathcal{I},\bW)=[f(\mathcal{I},\bw_1),\dots,f(\mathcal{I}, \bw_r)]$. After the deep architecture is trained, the hashing code for a new image $\mathcal{I}$ can be done by a simple quantization $b=sign(f(\mathcal{I}, \bW))$, where $sign(v)$ is a sign function on vectors that for $i=1,2,\dots,r$, $sign(v_i)=1$ if $v_i>0$, otherwise $sign(v_i)=0$.

\subsection{Structured Loss Optimization}

In deep metric learning for person re-identification, the network is often trained on data in the form of pairs \cite{FPNN,JointRe-id,DeepReID} or triplet ranking \cite{Ding:DeepRank}.  Thus, there are two commonly used cost functions, contrastive/pairwise loss and triplet ranking loss, which can be used in hash code optimization. We briefly revisit the two loss functions and then introduce the proposed structured loss function.

\paragraph{Contrastive/Pairwise Loss Function}
Given a person's binary codes $\bx_i$ and its correct match's codes $\by_i$ from a different camera view, the contrastive training tries to minimize the Hamming distance between a positive pair of $(\bx_i, \by_i)$ and penalize the negative pairs $(\bx_i,\by_j)$ ($i\neq j$) with a Hamming distance smaller than a margin. The contrastive cost function can be defined as
\begin{equation}
F = \sum_{(i,j)} a_{i,j} ||\bx_i - \by_j||_H + (1-a_{ij}) \left(\max\left(0,1 - ||\bx_i - \by_j||_H\right)\right)
\end{equation}
where $\bx_i$, $\by_j$ $\in \{0,1\}^r$ and $||\cdot||_H$ represents the Hamming distance. The label $a_{ij}\in \{0,1\}$ indicates whether a pair of binary codes $(\bx_i,\by_j)$ depicting the same person.

\paragraph{Triplet Ranking Loss Function}

Some recent studies have been made to learn hash functions that preserve relative similarities of the form of a triplet data $\left(\mathcal{I}, \mathcal{I}^+, \mathcal{I}^-\right)$ where image $\mathcal{I}$ (\textit{anchor}) of a specific person is more similar to all other images $\mathcal{I}^+$(\textit{positive}) of the same person than it is to any image $\mathcal{I}^-$ (\textit{negative}) of any other person (images $\mathcal{I}^+$ and $\mathcal{I}^-$ are from a different camera view from $\mathcal{I}$). Specifically, in hash function learning, the goad is to find a mapping $f(\cdot)$ such that the binary code $f(\mathcal{I})=\bx_i$ is closer to $f(\mathcal{I}^+)=\by_i$ than to $f(\mathcal{I}^-)=\by_j$ ($j\neq i$). Thus, we want
\begin{equation}\label{eq:triplet_constraint}
||\bx_i-\by_i||_H + 1 < ||\bx_i-\by_j||_H, \forall (\bx_i, \by_i, \by_j) \in \mathbb{T},
\end{equation}
where $\mathbb{T}$ is the set of all possible triplets in the training set and has cardinality $N$.
Accordingly, the triplet ranking hinge loss is defined by
\begin{equation}
\begin{aligned}
&F=\sum_i \max\left( 0, 1-\left( ||\bx_i -\by_j||_H - ||\bx_i-\by_i||_H \right)\right)\\
&s.t. ~~~ \bx_i, \by_i, \by_j \in \{0,1\}^r.
\end{aligned}
\end{equation}
A noticeable difference between a contrastive embedding and a triplet embedding is that a triplet unit with similar and dissimilar inputs provide some distance comparison context for the optimization process, as opposed to the contrastive loss that the network minimizes (same class) or maximizes (different classes) as much as possible for each pair independently \cite{TripletNet}.

In triplet embedding, however, generating all possible triplets would result in many triplets that easily fulfill the constraint in Eq \eqref{eq:triplet_constraint}, which is known as over-sampling. These triplets would not contribute to the training whereas resulting in slow convergence. An alternative strategy is to perform a smart sampling where one must be careful to avoid too much focus on hard training exemplars due to the possibility of over-fitting. Thus, it is crucial to \textit{actively}  select informative hard exemplars in order to improve the model.

Below, we introduce our structured loss which can avoid aforementioned over or under-sampling dilemmas by virtue of actively adding difficult neighbors to positive pairs into training batches.

\paragraph{The Proposed Structured Loss Function}
Previous works on person re-identification implement a Stochastic Gradient Decent (SGD) \cite{SGD} by drawing pairs or triplets of images uniformly at random. They didn't fully makes use of the information of the mini-batch that is sampled at a time and not only individual pairs or triplets. By contrast, we propose a structured loss over a mini-batch in order to take fully advantage of the training batches used in SGD. Meanwhile, the proposed structured loss can ensure fast convergence and stableness in training.

As shown in Fig.\ref{fig:loss} (c), the structured loss is conducted on all positive pairs and corresponding close (``difficult") negative pairs across camera views. Specifically, it can be formulated as
\begin{equation}\label{eq:structured_loss}
\begin{aligned}
&F = \frac{1}{|\bar{P}|} \sum_{\bx_i,\by_i \in \bar{P}} \max\left( 0, F_{\bx_i,\by_i}\right),\\
&F_{\bx_i,\by_i } = \max\left( \max\left(0, 1 - ||\bx_i-\by_k||_H\right), \max\left( 0, 1-||\by_i -\by_l||_H\right)\right)\\
& + ||\bx_i-\by_i||_H,  \\
&s.t. ~~ \bx_i,\by_i,\by_k,\by_l \in \{0,1\}^r, (\bx_i,\by_k)\in \bar{N}, (\by_i,\by_l)\in \bar{N},
\end{aligned}
\end{equation}
where $\bar{P}$ and $\bar{N}$ denote the set of positive and negative pairs in each mini-batch.  The process of selecting positive and negative samples is elaborated in Section \ref{ssec:hard_negative}.

\indent \textbf{Difference to contrastive and triplet ranking loss:}
\begin{itemize}
\item In pairwise training with $O(m)$ separate pairs in the batch,  a total of $O(m^2)$ pairs can be generated accordingly. However, these negative edges induced between randomly sampled pairs carry very limited information \cite{LiftEmbed}. By contrast, selected difficult exemplars are sharper cases that a full sub-gradient method would more likely focus on;
\item Compared with triplet embedding containing randomly sampled triplets, our training batch is augmented by adding negative neighbors bilaterally for each positive pairs. By doing this, the optimization process is conducted on most violate constraints, leading to fast convergence.
\end{itemize}

Fig.\ref{fig:loss} (a) and (b) illustrates a batch of positive/negative pairs and triplets with corresponding contrastive loss and triplet ranking loss. Green edges represent positive pairs (the same person) and red edges represent negative pairs (different individuals). Please note that these pairs and triplets are sampled completely random into a mini-batch. Fig.\ref{fig:loss} (c) illustrates the mining process for two positive pairs in the batch where for each image in a positive pair we seek its close (hard) negative images. We can see that our method allows mining the hard negatives from both the query image (\eg $\bx_1$ ) and its correct match (\eg $\by_1$) of a pair against gallery images (\eg $\by_m, m \neq 1$).

\paragraph{Optimization} For ease of optimization, we relax Eq.\eqref{eq:structured_loss} by replacing the Hamming norm with the $\ell_2$-norm and replacing the integer constraints on $\bx$'s and $\by$'s with the range constraints. The modified loss function is

\begin{equation}\label{eq:new_loss}
\begin{aligned}
&F = \frac{1}{|\bar{P}|} \sum_{\bx_i,\by_i \in \bar{P}} \max\left( 0, F_{\bx_i,\by_i}\right),\\
&F_{\bx_i,\by_i } = \max\left( \max\left(0, 1 - ||\bx_i-\by_k||_2^2\right), \max\left( 0, 1-||\by_i -\by_l||_2^2\right)\right)\\
& + ||\bx_i-\by_i||_2^2,  \\
& s.t. ~~ \bx_i,\by_i,\by_k,\by_l \in [0,1]^r, (\bx_i,\by_k)\in \bar{N}, (\by_i,\by_l)\in \bar{N}.
\end{aligned}
\end{equation}

The variant of structured loss is convex. Its sub-gradients with respect to $\bx_i$, $\by_i$, $\by_k$, and $\by_l$ are
\begin{equation}
\begin{aligned}
&\frac{\partial F }{\partial \bx_i}= (2\by_k-2\by_i) \times \mathbb{I}[2+ ||\bx_i-\by_i||_2^2 > ||\bx_i-\by_k||_2^2+||\by_i -\by_l||_2^2 ]\\
&\frac{\partial F}{\partial \by_i}=(2\by_l-2\bx_i) \times \mathbb{I}[2+ ||\bx_i-\by_i||_2^2 > ||\bx_i-\by_k||_2^2+||\by_i -\by_l||_2^2]\\
&\frac{\partial F}{\partial \by_k}=2\bx_i \times \mathbb{I}[2+ ||\bx_i-\by_i||_2^2 > ||\bx_i-\by_k||_2^2+||\by_i -\by_l||_2^2]\\
& \frac{\partial F}{\partial \by_l}=2\by_i \times \mathbb{I}[2+ ||\bx_i-\by_i||_2^2 > ||\bx_i-\by_k||_2^2+||\by_i -\by_l||_2^2]
\end{aligned}
\end{equation}
The indicator function $\mathbb{I}[\cdot]$ is the indicator function which outputs 1 if the expression evaluates to true and outputs 0 otherwise. Thus, the loss function in Eq.\eqref{eq:structured_loss} can be easily integrated into back propagation of neural networks. We can see that our structured loss provides informative gradients signals for all negative pairs which are within the margin of any positive pairs. In contrast to existing networks like \cite{FPNN,JointRe-id} where only hardest negative gradients are updated, making the training easily over-fit, the proposed structured loss makes the optimization much more stable.

\subsection{Hard Negative Mining for Mini-batches}\label{ssec:hard_negative}

As mentioned before, our approach differs from existing deep methods by making full information of the mini-batch that is sampled at a time, including positive pairs and their difficult neighbors. Please note that difficult neighbors are defined only with respect to the gallery camera view.
The motivation of doing this is to enhance the mini-batch optimization in network training because the sub-gradient of $F_{\bx_i,\by_i }$ would use the close negative pairs. Thus, our approach biases the sample towards including ``difficult" pairs.

In this paper, we particularly select a few positive pairs at random, and then \textbf{actively} add their difficult (hard) neighbors into the training mini-batch. This augmentation adds relevant information that a sub-gradient would use.  Specifically, we  determine the elements in mini-batches by online generation where all anchor-positive pairs in any identity are kept while selecting the hard negatives for both the anchor and its positive correspondence.
In fact, this procedure of mining hard negative edges amounts to computing the loss augmented inference in structured prediction setting \cite{Ioannis:icml2004,cutting-plane,LiftEmbed}. Intuitively, the loss from hard negative pairs should be penalized more heavily than a loss involving other pairs.
In this end, our structured loss function contains enough negative examples within the margin bound, which can push the positive examples towards the correct direction  and thus making the optimization much more stable.

\begin{figure}[t]
    \centering
        \includegraphics[width=4in,height=1.6in]{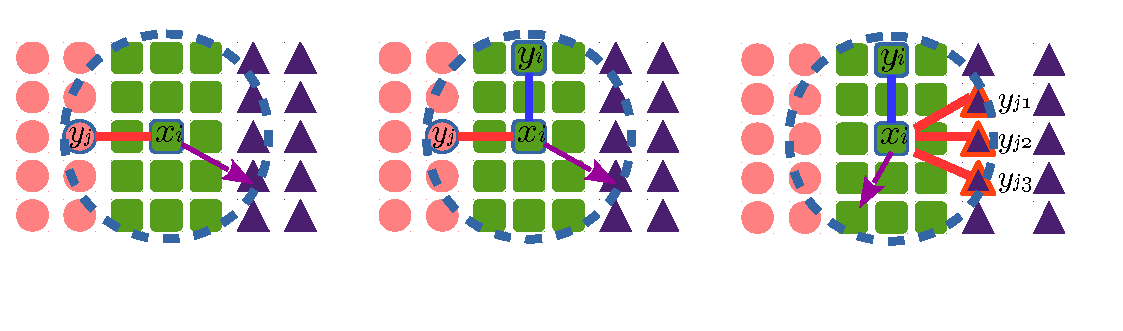}
    \caption{ Illustration on degeneration of contrastive and triplet ranking embedding with randomly sampled training pairs. Pink circles, green squares, and purple triangles indicate three different classes. Dotted blue circles regulate the margin bound where the loss becomes zero out of the bound. Magenta arrows denote the negative sub-gradient direction for positive samples. Left: Contrastive embedding. Middle: Triplet embedding. Right: Proposed structured embedding.}\label{fig:failure_mode}
\end{figure}

\begin{exam}
Fig.\ref{fig:failure_mode} shows failure cases in 2D profile with samples from three different classes, visualized by pink circles, green squares, and magenta triangles, respectively. The contrastive embedding has failure conditioned that randomly sampled negative $y_j$ is collinear \footnote{Three or more points are said to be collinear if they lie on a single straight line.} with examples from a third class (purple triangles). For triplet embedding, the degenerated case happens when a negative $y_j$ is within the margin bound with respect to the anchor $x_i$ and its positive $y_i$. In this situation, both contrastive and triplet embedding incorrectly enforce the gradient direction of positives towards examples from the third class. By contrast, through explicitly mining hard negatives within the margin w.r.t. the positive $x_i$, the proposed structured embedding can push the positives towards the correct direction.
\end{exam}

\begin{thm}
Margin maximization.
Hard negative mining on mini-batches is equivalent to computing the loss augmented inference, which promotes margin maximization in pairwise/triplet units.
\end{thm}
\emph{Proof}.
Following the definitions in Eq.\eqref{eq:structured_loss}, the condition of zero training error can be compactly written as a set of non-linear constraints
\begin{equation}\label{eq:nonlinear}
\forall i: \max_{\by\in \mathcal{Y} \char`\\ \by_i} \{\langle\bw, H(\bx_i,\by)\rangle\} < \langle\bw, H(\bx_i,\by_i)\rangle.
\end{equation}
where $\mathcal{Y}$ contains training samples from cross-camera view against $\bx_i$. $H(.)$ denotes Hamming distance.
Each non-linear inequality in Eq.\eqref{eq:nonlinear} can be equivalently replaced by $|\mathcal{Y}|-1$ linear inequalities, and thus we have
\begin{equation}\label{eq:margin}
\begin{split}
&\forall i, \forall \by \in \mathcal{Y} \char`\\ \by_i: \langle \bw, \delta H_i(\by)\rangle <0;\\
&\delta H_i(\by) \equiv H(\bx_i,\by)-H(\bx_i,\by_i).
\end{split}
\end{equation}
Recall Eq.\eqref{eq:structured_loss} that the hard negative mining is equivalent to augmenting the loss as $\bar{H}_i (\by)=H(\bx_i,\by)-H(\bx_i,\by_i)+H(\by_i,\by)$. Thus, the linear constraint in Eq.\eqref{eq:margin} is updated as
\begin{equation}\label{eq:augmented_margin}
\begin{split}
&\forall i, \forall \by \in \mathcal{Y} \char`\\ \by_i: \langle \bw, \delta \bar{H}_i(\by)\rangle <0;\\
& \Leftrightarrow \langle \bw, \delta H_i(\by)\rangle + \langle \bw, H(\by_i,\by) \rangle<0.
\end{split}
\end{equation}
In Eq.\eqref{eq:augmented_margin}, since the term $\langle \bw, H(\by_i,\by) \rangle \geq 1-\epsilon_i$, $\epsilon_i\geq 0$ is a small slack variable, the term $\langle \bw, \delta H_i(\by)\rangle$ is imposed a more tight constraint on its margin maximization. $\Box$

\begin{figure*}[ht]
 \hspace{-1cm} \includegraphics[width=1.8in,height=1.5in]{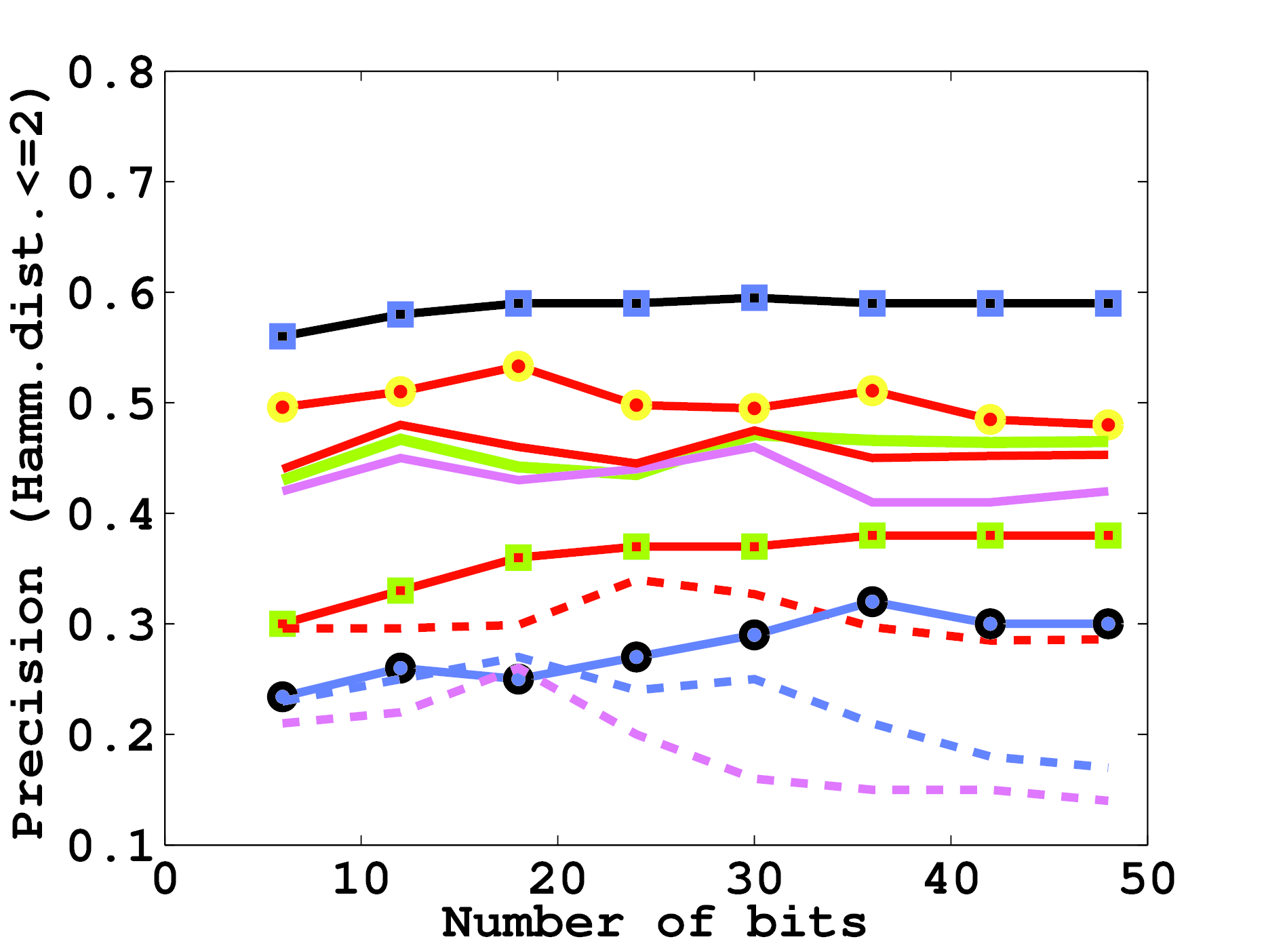}
  \hspace{-0.5cm} \includegraphics[width=1.8in,height=1.5in]{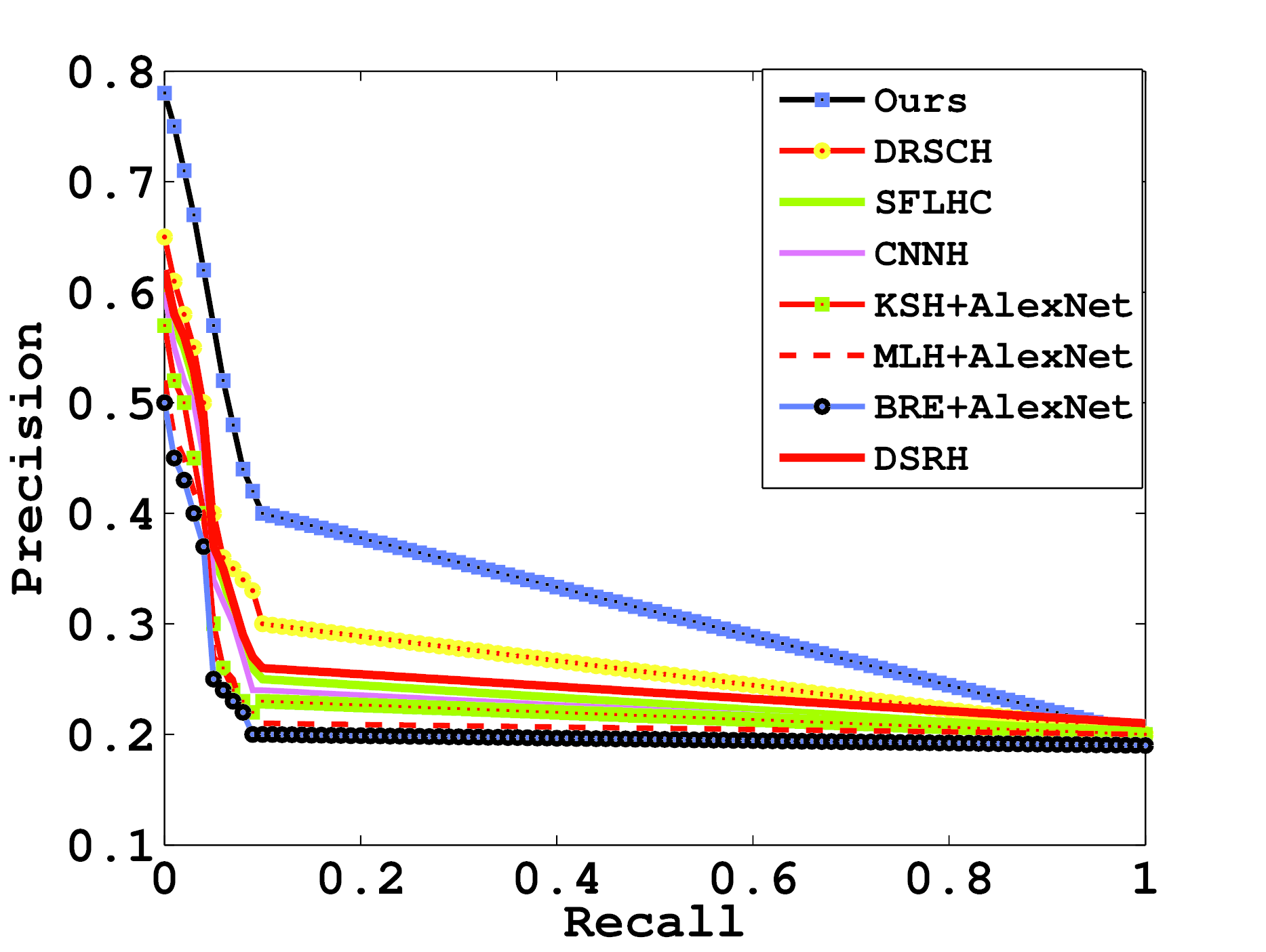}
\hspace{-0.5cm}	\includegraphics[width=1.8in,height=1.5in]{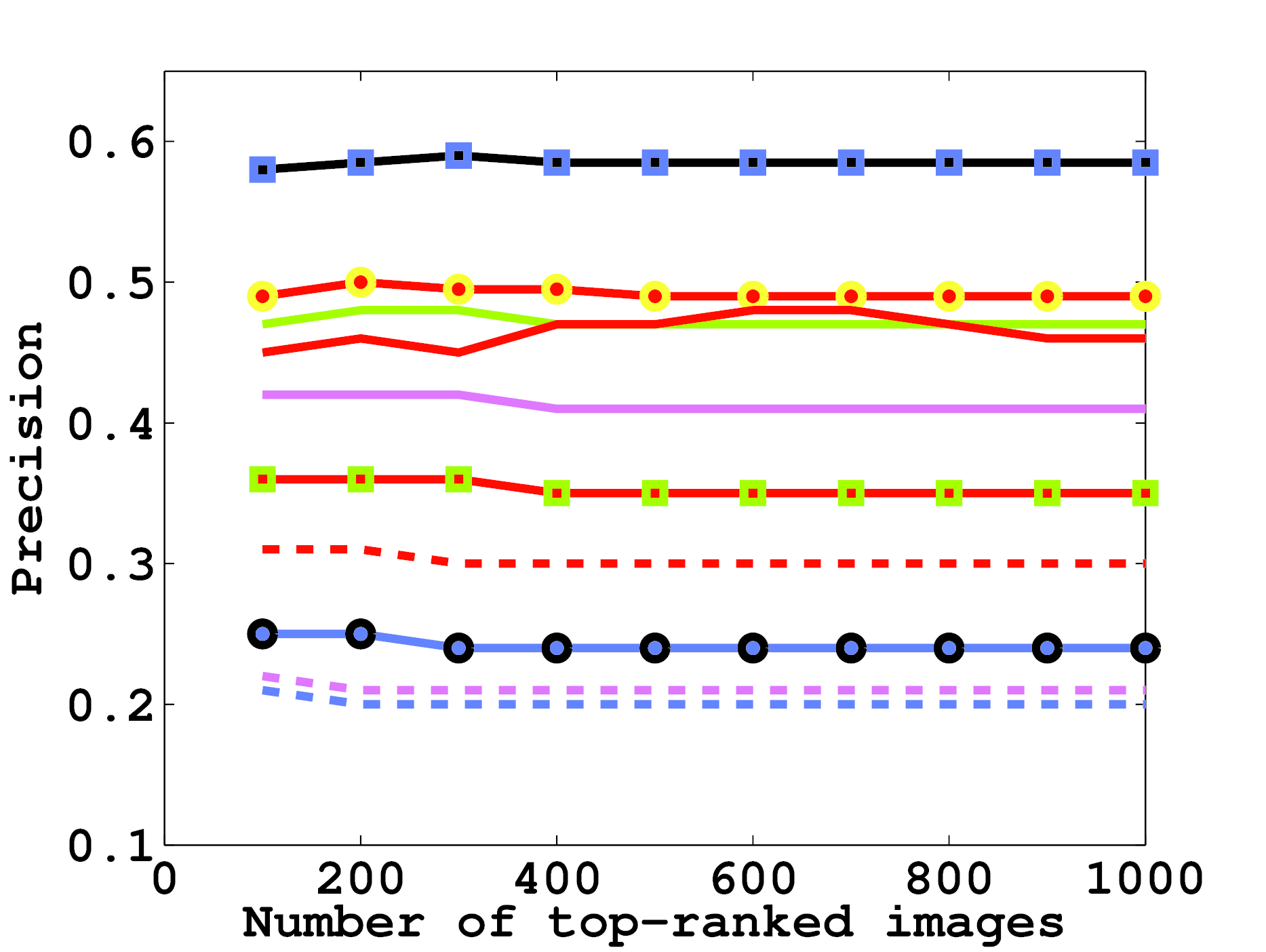}\\
\hspace{5cm}(a) \hspace{4cm} (b) \hspace{4cm} (c)\\
    \caption{The results on CUHK03. (a) precision curves within Hamming radius; (b) precision-recall curves of Hamming ranking with 48 bits; (c) precision curves with 48 bits with respect to varied number of top-ranked images. Best view in color.}\label{fig:results_cuhk03}
\end{figure*}

\begin{figure*}[ht]
  \hspace{-1cm}   \includegraphics[width=1.8in,height=1.5in]{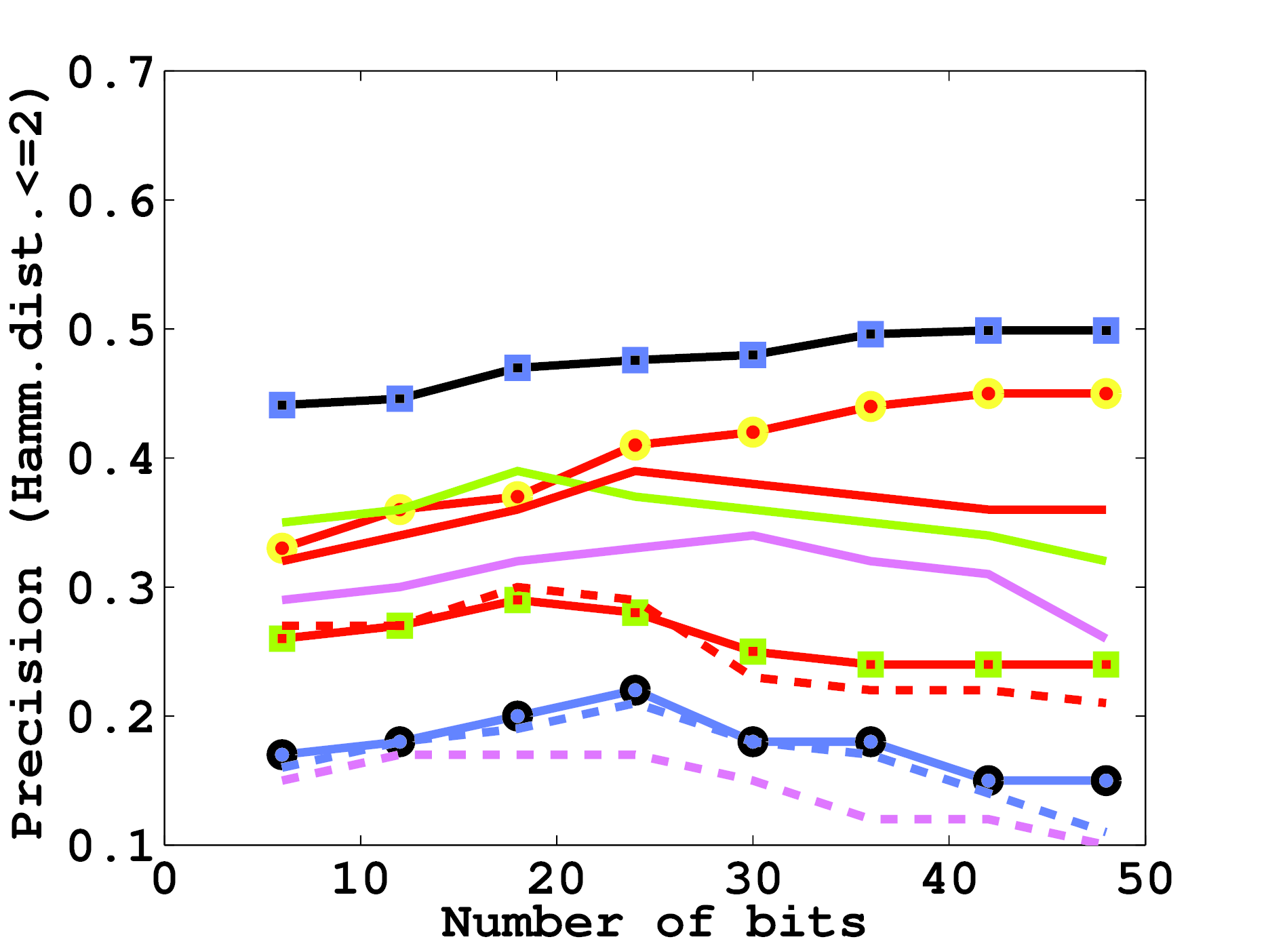}
    \hspace{-0.5cm}     \includegraphics[width=1.8in,height=1.5in]{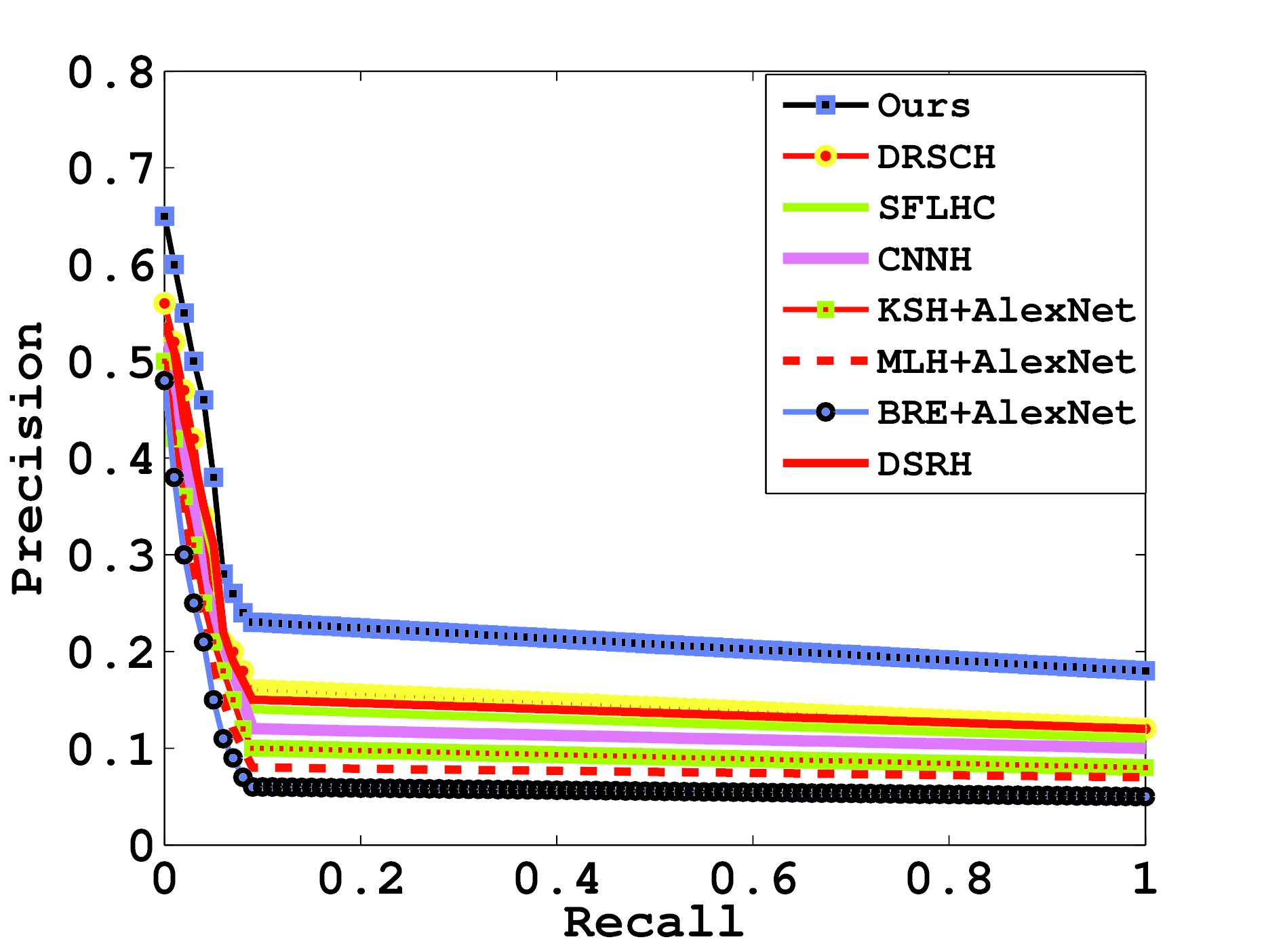}
\hspace{-0.5cm}	\includegraphics[width=1.8in,height=1.5in]{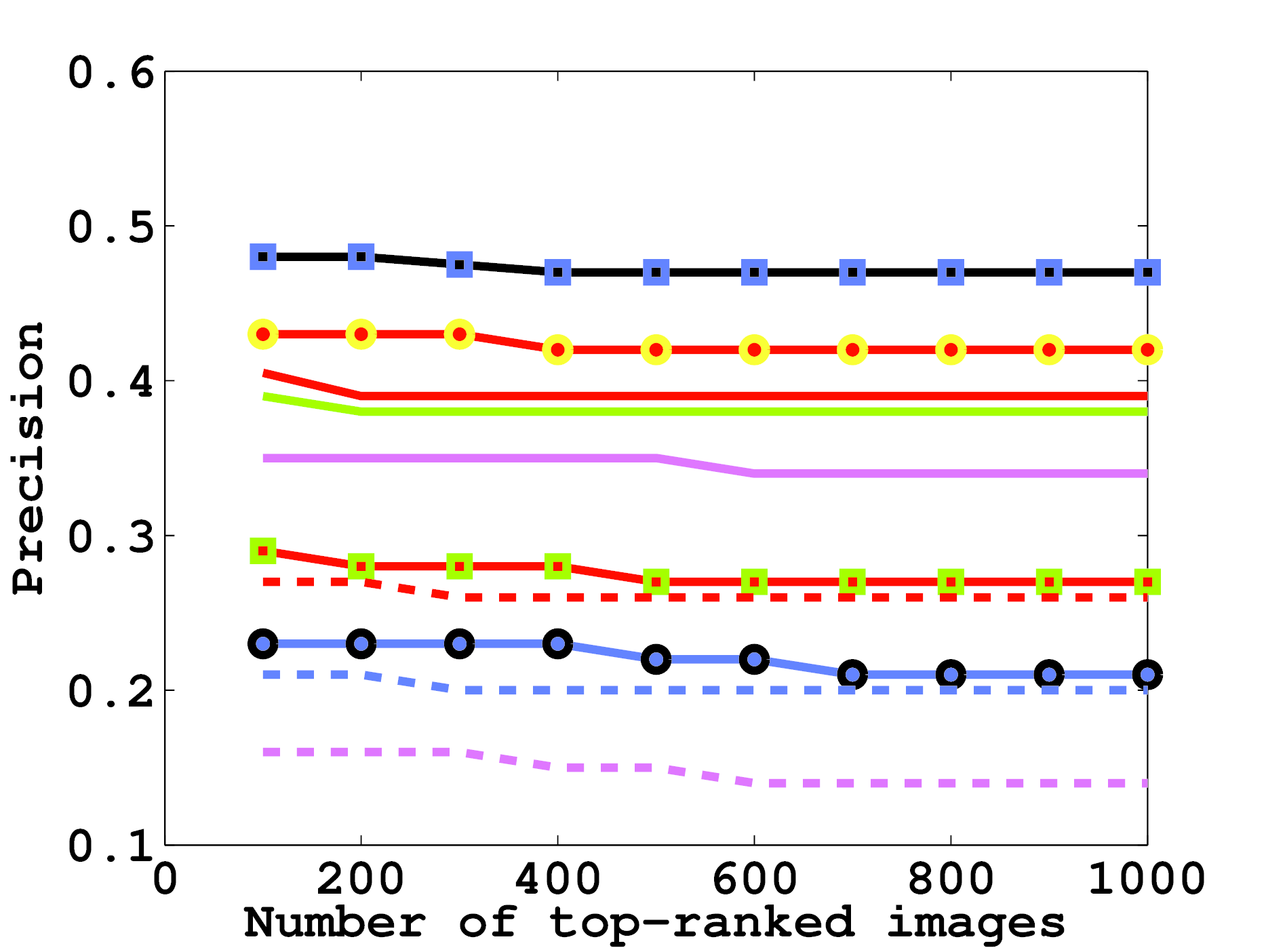}\\
\hspace{5cm}(a) \hspace{4cm} (b) \hspace{4cm} (c)\\
    \caption{The results on Market-1501. (a) precision curves within Hamming radius; (b) precision-recall curves of Hamming ranking with 48 bits; (c) precision curves with 48 bits with respect to varied number of top-ranked images. Best view in color.}\label{fig:results_market}
\end{figure*}

\begin{table*}[hbt]
\caption{MAP of Hamming ranking w.r.t. varied number of bits on two person re-identification datasets.}\label{tab:MAP}
\centering
\begin{tabular}{|l|c|c|c|c|c|c|c|c|}
\hline\hline
Method& \multicolumn{4}{c|}{CUHK03} & \multicolumn{4}{c|}{Market-1501}\\
\cline{2-9}
& 24 bits & 32 bits & 48 bits & 128 bits & 24 bits & 32 bits & 48 bits & 128 bits\\
\hline
Ours & \textbf{0.579} & \textbf{0.594} & \textbf{0.602} & \textbf{0.601} &\textbf{0.452} & \textbf{0.466} & \textbf{0.481} & \textbf{0.482}\\
SFLHC \cite{Lai:hash-deep}  & 0.428 & 0.468 & 0.472 & 0.476  & 0.365 & 0.372 & 0.377  & 0.378\\
DSRH \cite{DeepSemanticHash} &0.494 & 0.486& 0.482&  0.484 & 0.403 & 0.421 &0.426  & 0.423\\
DRSCH \cite{DeepRegularlizedHash} & 0.509 & 0.517 & 0.523 & 0.521 & 0.421 & 0.443 & 0.452 & 0.447\\
\hline
CNNH \cite{Xia:hash2014} & 0.403 & 0.411 & 0.417 & 0.414 &0.305 & 0.337 & 0.346 & 0.345 \\
KSH+AlexNet \cite{Liu:kernel-hash} & 0.301 & 0.339 & 0.356 & 0.357  & 0.264 &0.282  & 0.288 & 0.281\\
MLH+AlexNet \cite{Norouzi:loss-hash} & 0.262 & 0.295 & 0.299 & 0.302  & 0.224 & 0.257 & 0.269  & 0.273\\
BRE+AlexNet \cite{Kulis:nips2009} &0.206 & 0.215 & 0.237 & 0.239 & 0.185 & 0.196  & 0.211 & 0.210 \\
\hline
\end{tabular}
\end{table*}

\section{Experiments}\label{sec:exp}

In this section, we conduct extensive evaluations of the proposed architecture on two largest datasets in person re-identification: \textbf{CUHK03} \cite{FPNN} and \textbf{Market-1501} \cite{Market1501}.

\subsection{Experimental Settings}

\paragraph{Datasets} Person re-identification comes with a number of benchmark datasets such as VIPeR \cite{Gray2007Evaluating}, PRID2011 \cite{Hirzer2011Person}, and iLIDS \cite{Zheng2009Associating}. However, these datasets are moderately small/medium-sized, rendering them not suitable to be the test bed for our fast hashing learning framework. More recently, to facilitate deep learning in person re-identification, two large datasets \ie CUHK03 and Market1501 are contributed with more identities captured by multiple cameras in more realistic conditions.

\begin{itemize}
\item The \textbf{CUHK03} dataset \cite{FPNN} includes 13,164 images of 1360 pedestrians. The whole dataset is captured with six surveillance camera. Each identity is observed by two disjoint camera views, yielding an average 4.8 images in each view. This dataset provides both manually labeled pedestrian bounding boxes and bounding boxes automatically obtained by running a pedestrian detector \cite{DetectionPAMI}. In our experiment, we report results on labeled data set.

\item The \textbf{Market-1501} dataset \cite{Market1501} contains 32,643 fully annotated boxes of 1501 pedestrians, making it the largest person re-id dataset to date. Each identity is captured by at most six cameras and boxes of person are obtained by running a state-of-the-art detector, the Deformable Part Model (DPM) \cite{MarketDetector}.  The dataset is randomly divided into training and testing sets, containing 750 and 751  identities, respectively.
\end{itemize}

\paragraph{Competitors} We present quantitative evaluations in terms of searching accuracies and compare our method with seven state-of-the-art methods:
\begin{itemize}
\item Kernel-based Supervised Hashing (KSH) \cite{Liu:kernel-hash}: KSH is a kernel based method that maps the data to binary hash codes by maximizing the separability of code inner products between similar and dissimilar pairs. In particular, KSH adopts the kernel trick to learn nonlinear hash functions on the feature space.
\item Minimal Loss Hashing (MLH) \cite{Norouzi:loss-hash}: MLS is working by treating the hash codes ad latent variables, and employs the structured prediction formulation for hash learning.
\item Binary Reconstructive Embedding (BRE) \cite{Kulis:nips2009}: Without requiring any assumptions on data distributions, BRE directly learns the hash functions by minimizing the reconstruction error between the distances in the original feature space and the Hamming distances in the embedded binary space.
\item CNNH \cite{Xia:hash2014}: is a supervised hashing method in which the learning process is decomposed into a stage of learning approximate hash codes, followed by a second stage of learning hashing functions and image representations from approximate ones.
\item Simulaneous Feature Learning and Hash Coding based on CNNs (SFLHC) \cite{Lai:hash-deep}: SFLHC is a deep architecture consisting of stacked convolution layers and hashing code learning module. It adopts a triplet ranking loss to preserve relative similarities.
\item Deep Semantic Ranking Hashing (DSRH) \cite{DeepSemanticHash}: DSRH is a recently developed method that incorporates deep feature learning into hash framework in order to preserve multi-level semantic similarity between multi-label images. Also, their network is optimized on a triplet ranking embedding.
\item Deep Regularized Similarity Comparison Hashing (DRSCH) \cite{DeepRegularlizedHash}: DRSCH is a deep framework which aims to generate bit-scalabel hash codes directly from raw images. Their network is optimized by triplet ranking loss, and hash codes are regularized by adjacency consistency.
\end{itemize}

The first three methods are conventional supervised methods and the last three are based on deep learning framework. The results of these competitors are obtained by the implementations provided by their authors.
For fair comparison on three supervised methods \ie KSH, MLH, and BRE, we extract CNN features for person images using AlexNet \cite{Krizhevsky2012Imagenet}, and feed the feature vectors from the last fully-connected layer (4096-dim) to MLH and BRE, denoted as KSH+AlexNet, MLH+AlexNet, BRE+AlexNet, respectively.

\begin{figure}[t]
    \centering
        \includegraphics[width=3in,height=1.2in]{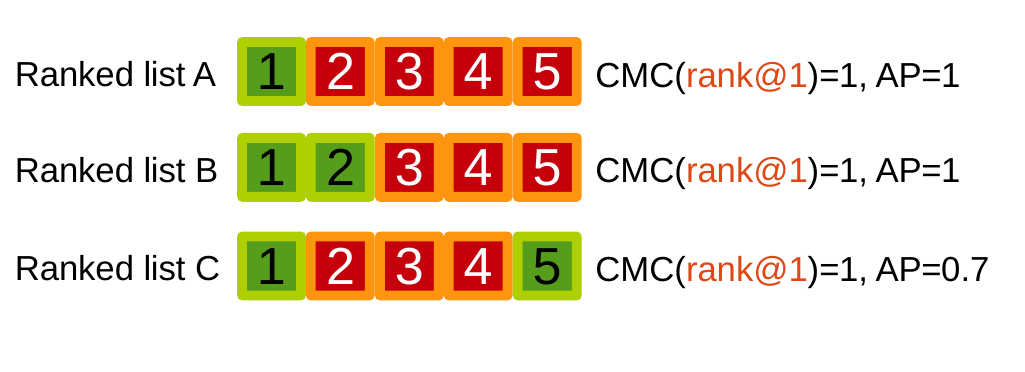}
    \caption{The difference between Average Precision (AP) and CMC measurements. The green and red boxes represent the position of true and false matches in rank lists. For all three rank lists, CMC curve at rank 1 remains 1 whilst AP=1 (rank list A), 1 (rank list B), and 0.7 (list C), respectively.}\label{fig:AP_CMC}
\end{figure}

\paragraph{Evaluation Protocol} We adopt four evaluation metrics in the experiments: Mean Average Precision (MAP), Precision curves with Hamming distance within 2, Precision-Recall curves, and Precision curves with respect to varied number of top returned samples.

In person re-identification, a standard evaluation metric is Cumulated Matching Characteristics (CMC) curve, which shows the probability that a correct match to the query identity appears in different-sized candidate lists. This measurement is, however, is valid only in the single-shot setting where there is only one ground truth match for a given query (see an example in Fig.\ref{fig:AP_CMC}). In the case of one-shot, precision and recall are degenerated to be the same manner. Nonetheless, given multiple ground truths regarding to a query identity, the CMC curve is biased due to the fact that the recall issue is not considered. For instance, two rank lists A and B in Fig.\ref{fig:AP_CMC} can yield their CMC value equal to 1 at rank=1, respectively, whereas CMC curves fail to provide a fair comparison of the quality between the two rank lists. By contrast, Average Precision (AP) can quantitatively evaluate the quality of rank list for the case of multi-ground-truth.

For Market-1501 (CUHK03) dataset, there are on average 14.8 (4.8) cross-camera ground truths for each query. Thus, we employ Mean Average Precision (MAP) to evaluate the overall performance. For each query, we calculate the area under the Precision-Recall curve, which is known as Average Precision (AP). Then, MAP is calculated as the mean value of APs over all queries. We have the definition of MAP in the following
\begin{equation}\label{eq:MAP}
MAP(Q)=\frac{1}{Q}\sum_{j=1}^{|Q|}\frac{1}{m_j}\sum_{k=1}^{m_j}Precision(R_{jk}),
\end{equation}
where $Q$ denotes a set of queries, and $\{d_1,\dots,d_{mj}\}$ are a set of relevant items with respect to a given query $q_j\in Q$. $R_{jk}$ is the set of ranked retrieval results from the top results until item $d_k$ is retrieved.

Given a query, the precision with hamming distance within 2 (@ $r$-bits) w.r.t. the returned top $N$ nearest neighbors is defined as
\begin{equation}\small
Precision (||\cdot||_H<=2) @ N =\frac{\sharp \left(imgs \cap ||imgs-query||_H<=2\right)}{N}
\end{equation}
where $imgs$ denote similar images to the query, the hamming distance between two binary vectors is the number of coefficients where they differ.
The four types of metrics are widely used to evaluate hashing models \cite{Liu:kernel-hash,Lai:hash-deep}.

\paragraph{Implementation Details}  We implemented our architecture using the Theano \cite{Theano} deep learning framework with contrastive, triplet, and the proposed structured loss. The batch size is set to 128 for contrastive and our method and to 120 for triplet. Network training converges in roughly 22-24 hours on NVIDIA GTX980. All training and test images are normalized to 160 by 60. We augment the training data by performing random 2D translation, as also done in \cite{FPNN,JointRe-id}. For an original image of size $W\times H$, we sample 5 images around the image center, with translation drawn from a uniform distribution in the range $[-0.05H,0.05H]\times [-0.05W,0.05W]$. In training, we exhaustively use all the positive pairs of examples and randomly generate approximately equal number of negative pairs as positives.

In Market-1501, there are 12,936 images for training and 19,732 images for test, corresponding to 750 and 751 identities, respectively.  In CUHK03 dataset, we randomly partition the dataset into training, validation, and test with 1160, 100, and 100 identities, respectively. During testing, for each identity, we select one query image in each camera. The search process is performed in a \emph{cross-camera} mode, that is, relevant images captured in the same camera as the query are regarded as ``junk" \cite{ObjRetrieval}, which means that this image has no influence to re-identification accuracy. In this scenario, for Market-1501 dataset, each identity has at most 6 queries, and there are 3,363 query images in total. For CUHK03 dataset, each identity has at most 2 queries, and there are 200 query images in total.

In our implementation, we use all positive anchor positive pairs regarding to each identity. In pairwise training, anchor negative pairs are generated by randomly selecting a sample from a different identity with respect to the anchor's identity. The same sampling scheme is applied on triplet selection. To add meaningful hard negatives into mini-batch in our model, we select hard neighbors from gallery view for each training image in a positive pair. Specifically, for an anchor $\mathcal{I}$ and its positive $\mathcal{I}^+$, their hard negatives $\mathcal{I}^-$s  are selected such that $||s_{\mathcal{I} } - s_{\mathcal{I}^+}||_2^2 < ||s_{\mathcal{I} } - s_{\mathcal{I}^-} ||_2^2$, where $s_{ (\cdot) }$ is a visual descriptor and in our experiment we use SIFT features at the beginning of training \footnote{To extract SIFT/LAB features, we first rescale pedestrian images to a resolution of $48\times 128$ in order to remove some background regions. SIFT and color features are extracted over a set of 14 dense overlapping $32\times 32$-pixels regions with a step stride of 16 pixels in both directions. Thus, we have the feature descriptor vector with length 9408.}. Since features are updated continuously as network is on training, $s_{ (\cdot) }$ corresponds to feature extracted after each 50 epochs.

\begin{table}[ht]
\caption{Comparison results of skip layer (FC1+FC2) against second fully connected layer (FC2) on two datasets of CUHK03 and Market-1501.}\label{tab:FC_layer}
\centering
\begin{tabular}{|c|c|c|c|c|}
    \hline\hline
   Method & 24 bits & 32 bits & 48 bits & 128 bits\\
  \hline
    \multicolumn{5}{|c|}{MAP (CUHK03)}\\
    \cline{1-5}
    FC2 & 0.529 & 0.546 & 0.571 & 0.584\\
    FC1+FC2 & \textbf{0.579} &  \textbf{0.594} & \textbf{0.602} &\textbf{0.601} \\
    \hline
    \multicolumn{4}{|c|}{MAP (Market-1501)}\\
    \cline{1-5}
    FC2 & 0.417 &  0.420 & 0.439 & 0.437 \\
    FC1+FC2 & \textbf{0.452} & \textbf{0.466} & \textbf{0.481} & \textbf{0.482} \\
    \hline
\end{tabular}
\end{table}

\subsection{Results on Benchmark Datasets}

We test and compare the search accuracies of all methods against two datasets. Comparison results are reported in Table \ref{tab:MAP} and Figs. \ref{fig:results_cuhk03}--\ref{fig:results_market}. We can see that
\begin{itemize}
\item On the two benchmark datasets, the proposed method outperforms all supervised learning baselines using CNN features in terms of MAP, precision with Hamming distance 2, precision-recall, and precision with varying size of top returned images. For instance, compared with KSH + AlexNet, the MAP results of the proposed method achives a gain from 35.6\%~58.5\%, 28.8\%~48.1\% with 48 bits on CUHK03 and Market-1501, respectively.
\item Comparing with  CNNH \cite{Xia:hash2014}, which is a two-stage deep network based hashing method, our method indicates a better searching accuracies.  Specifically, the MAP results achieve a relative increase by a margin of 16\% and 13\% on two datasets, respectively. This observation can verify that jointly learning features and hashing codes are beneficial to each other.
\item Comparing with the most related competitors DSRH \cite{DeepSemanticHash} and DRSCH \cite{DeepRegularlizedHash}, our structured prediction suits well to SGD and thus achieves superior performance. For example, in terms of MAP on CUHK03 dataset, a notable improvement can be seen from 49.4\% (50.9\%) to 54.7\%, compared with  DSRH \cite{DeepSemanticHash} (DRSCH \cite{DeepRegularlizedHash}).
\end{itemize}

We also conduct self-evaluation of our architecture with skip layer connected to hash layers and its alternative with only the second fully connected layer. As can be seen in Table \ref{tab:FC_layer}, the results of the proposed architecture outperforms its alternative with only the second fully connected layer as input to the hash layer. One possible reason is the hash layer can see multi-scale features by connecting to the first and second fully connected layers (features in the FC2 is more global than those in FC1). And adding this bypass connections can reduce the possible information loss in the network.

\subsection{Comparison with State-of-the-art Approaches}
In this section, we evaluate our method by comparing with state-of-the-art approaches in person re-identification.
Apart from the above hashing methods, seven competitors are included in our experiment, which are  FPNN \cite{FPNN}, JointRe-id \cite{JointRe-id}, KISSSME \cite{KISSME}, SDALF \cite{Farenzena2010Person}, eSDC \cite{eSDC}, kLFDA \cite{Xiong2014Person}, XQDA \cite{LOMOMetric}, DomainDropout \cite{DomainDropout}, NullSpace \cite{NullSpace-Reid} and BoW \cite{Market1501}. For KISSME \cite{KISSME}, SDALF \cite{Farenzena2010Person}, eSDC \cite{eSDC}, kLFDA \cite{Xiong2014Person} and BoW model \cite{Market1501}, the experimental results are generated by their suggested features and parameter settings. For XQDA \cite{LOMOMetric} and NullSpace \cite{NullSpace-Reid},  the Local Maximal Occurrence (LOMO) features are used for person representation. The descriptor has 26,960 dimensions.  FPNN \cite{FPNN} is a deep learning method with the validation set adopted to select parameters of the network. JointRe-id \cite{JointRe-id} is an improved deep learning architecture in an attempt to simultaneously learn features and a corresponding similarity metric for person re-identification.
DomainDropout \cite{DomainDropout} presents a framework for learning deep feature representations from multiple domains with CNNs. We also extract the intermediate features from the last fully-connected layer, denoted as Ours (FC), to evaluate the performance without hash layer. To have fair comparison with DomainDropout \cite{DomainDropout}, we particularly leverage training data from CUHK03, CUHK01 \cite{CUHK01} with domain-aware dropout, and Market-1501, denoted as Ours (DomainDropout).

Table \ref{tab:compare_state_art} displays comparison results with state-of-the-art approaches, where all of the Cumulative Matching Characteristics (CMC) Curves are single-shot results on CUHK03 dataset whilst multiple-shot on Market1501 dataset. All hashing methods perform using 128 bits hashing codes, and the ranking list is based on the Hamming distance.
We can see that on Market-1501 dataset our method outperforms all baselines on rank 1 recognition rate except NullSpace \cite{NullSpace-Reid}. The superiority of NullSpace \cite{NullSpace-Reid} on Market-1501 comes from enough samples in each identity, which allows it to learn a discriminative subspace. Our result (48.06\%) is very comparative to NullSpace \cite{NullSpace-Reid} (55.43\%) while the time cost is tremendously reduced, as shown in Table \ref{tab:test_time}. Besides, the performance of our model without hash layer (Ours (FC)) is consistently better than that with hashing projection. This is mainly because the dimension reduction in hashing layer and quantization bring about certain information loss.

On CUHK03 dataset, DomainDropout \cite{DomainDropout} achieves the best performance in recognition rate at rank from 1 to 10. This is mainly because DomainDropout \cite{DomainDropout} introduces a method to jointly utilize all datasets in person re-identification to produce generic feature representation. However, this action renders their model extremely expensive in training given a variety of datasets varied in size and distributions. To this end, we test the average testing time of our model and competing deep learning methods, and report results in Table \ref{tab:test_time}. The testing time aggregates computational cost in feature extraction, hash code generation, and image search. For all the experiments, we assume that every image in the database has been represented by its binary hash codes. In this manner, the time consumption of feature extraction and hash code generation is mainly caused by the query image.  It is obvious that our model achieves comparable performance in terms of efficiency in matching pedestrian images. Our framework runs slightly slower than DRSCH and SFLHC due to the computation of structured loss on each mini-batch.

\begin{table*}[hbt]
\caption{Comparison with state-of-the-art approaches on two person re-identification datasets. The evaluation is based on CMC. ``-" indicates that the result is not available.}\label{tab:compare_state_art}
\centering
\begin{tabular}{|l|c|c|c|c|c|c|c|c|}
\hline\hline
Method& \multicolumn{4}{c|}{CUHK03 (CMC \%)} & \multicolumn{4}{c|}{Market-1501 (CMC \%) }\\
\cline{2-9}
& r=1 & r=5 & r=10 & r=20 & r=1 & r=5 & r=10 & r=20\\
\hline
Ours & 37.41 & 61.28 & 77.46 & 88.42 & 48.06 & 61.23 & 75.67 & 87.06 \\
Ours (FC2) & 43.20 & 69.07 & 84.24 & 90.92 & 50.12 & \textbf{63.50} & \textbf{76.82} & \textbf{89.24}\\
Ours (DomainDropout) & \textbf{74.20} & \textbf{92.27} & 94.24 & 95.92 & \textbf{58.12} & \textbf{68.50} & \textbf{80.82} & \textbf{92.24}\\
SFLHC \cite{Lai:hash-deep}  & 12.38 & 30.52 & 49.34 & 71.55 & 37.74 & 59.09 & 74.25 & 86.52\\
DSRH \cite{DeepSemanticHash} & 9.75 & 28.10 & 47.82 & 67.95  & 34.33 & 59.82 & 71.27 & 86.09\\
DRSCH \cite{DeepRegularlizedHash} & 20.84 & 49.39 & 72.66 & 83.03 & 41.25 & 58.98 & 76.04 & 85.33 \\
\hline
CNNH \cite{Xia:hash2014} & 8.27 & 22.53 & 45.09 & 59.74 & 16.46 & 39.95 & 51.24 & 71.23\\
KSH+AlexNet \cite{Liu:kernel-hash} &  5.65 & 15.71 & 22.75 & 34.68  & 12.57 & 31.22 & 48.66 & 66.72\\
MLH+AlexNet \cite{Norouzi:loss-hash} &  5.75 & 15.62 & 27.61 &  42.68  & 10.89 & 29.93 & 46.78 & 66.32\\
BRE+AlexNet \cite{Kulis:nips2009} & 4.91 & 11.24 & 17.83 & 29.20  & 12.65 & 32.68 & 49.08 & 67.70\\\hline
FPNN \cite{FPNN} & 20.65 & 50.09 & 66.42 & 80.02  & - & -  & -  & -\\
JointRe-id \cite{JointRe-id} & 54.74 & 86.71 & 91.10 & \textbf{97.21} & - & -  & -  & -\\
KISSME \cite{KISSME} & 14.17 & 41.12  & 54.89 & 70.09  & 40.47 & 59.35  & 75.26 & 83.58\\
SDALF \cite{Farenzena2010Person} & 5.60 & 23.45  & 36.09 & 51.96 & 20.53 & 47.82 & 62.45 & 80.21\\
eSDC \cite{eSDC} & 8.76 & 27.03 &  38.32 & 55.06  & 33.54 & 58.25 & 74.33 & 84.57 \\
kLFDA \cite{Xiong2014Person} & 47.25 & 64.58 & 82.36 & 89.17  & -  &  - & -  & -\\
BoW \cite{Market1501} & 24.33 & 58.42 & 71.28 & 84.91  & 47.25 & 62.71  & 75.33  & 86.42\\
DomainDropout \cite{DomainDropout} & 72.60 & 92.22 & \textbf{94.50} & 95.01  & - & -  & -  & -\\
NullSpace (LOMO \cite{LOMOMetric}) \cite{NullSpace-Reid} & 58.90 & 85.60 & 92.45 & 96.30  & 55.43 & -  & -  & -\\
XQDA (LOMO) \cite{LOMOMetric} & 52.20 & 82.23 & 92.14 & 96.25  & 43.79 & -  & -  & -\\
\hline
\end{tabular}
\end{table*}

\begin{table}[hbt]
\tiny
\caption{Comparison on the average testing time (millisecond per image) by fixing the code length to be 48 (128) bits. $\infty$ indicates that the computational cost is too high to be estimated and ``-" indicates that the result is not available.}\label{tab:test_time}
\centering
\begin{tabular}{|l|c|c|}
\hline\hline
Method& CUHK03 (ms) & Market-1501 (ms) \\
\hline
Ours & 4.617 (4.982) & 7.374 (7.902) \\
SFLHC \cite{Lai:hash-deep}  & 4.241 (4.782)& 5.892 (6.417)\\
DSRH \cite{DeepSemanticHash} & 5.765 (6.019) & 7.887 (8.445) \\
DRSCH \cite{DeepRegularlizedHash} & 2.332 (2.816) & 3.609 (3.973) \\
\hline
CNNH \cite{Xia:hash2014} & 5.359 (5.743) & 6.943 (7.410)\\
KSH+AlexNet \cite{Liu:kernel-hash} &  7.279 (7.805) & 9.046 (9.537) \\
MLH+AlexNet \cite{Norouzi:loss-hash} & 6.727 (7.198)  & 8.092 (8.545)\\
BRE+AlexNet \cite{Kulis:nips2009} & 6.765 (7.214) & 9.072 (9.613)\\
FPNN \cite{FPNN} & $\infty$ & $\infty$  \\
JointRe-id \cite{JointRe-id} & $\infty$ & $\infty$ \\
DomainDropout \cite{DomainDropout} &$\infty$ &$\infty$\\
kLFDA \cite{Xiong2014Person} & - & $43.4\times 10^3$\\
XQDA \cite{LOMOMetric} & - & $1.6\times 10^3$\\
MFA \cite{Xiong2014Person} &- & $43.2\times 10^3$\\
NullSpace \cite{NullSpace-Reid} &- & $31.3\times 10^3$\\
\hline
\end{tabular}
\end{table}

\subsection{Convergence Study}

\begin{figure}[ht]
    \centering
        \includegraphics[width=1.7in,height=1.3in]{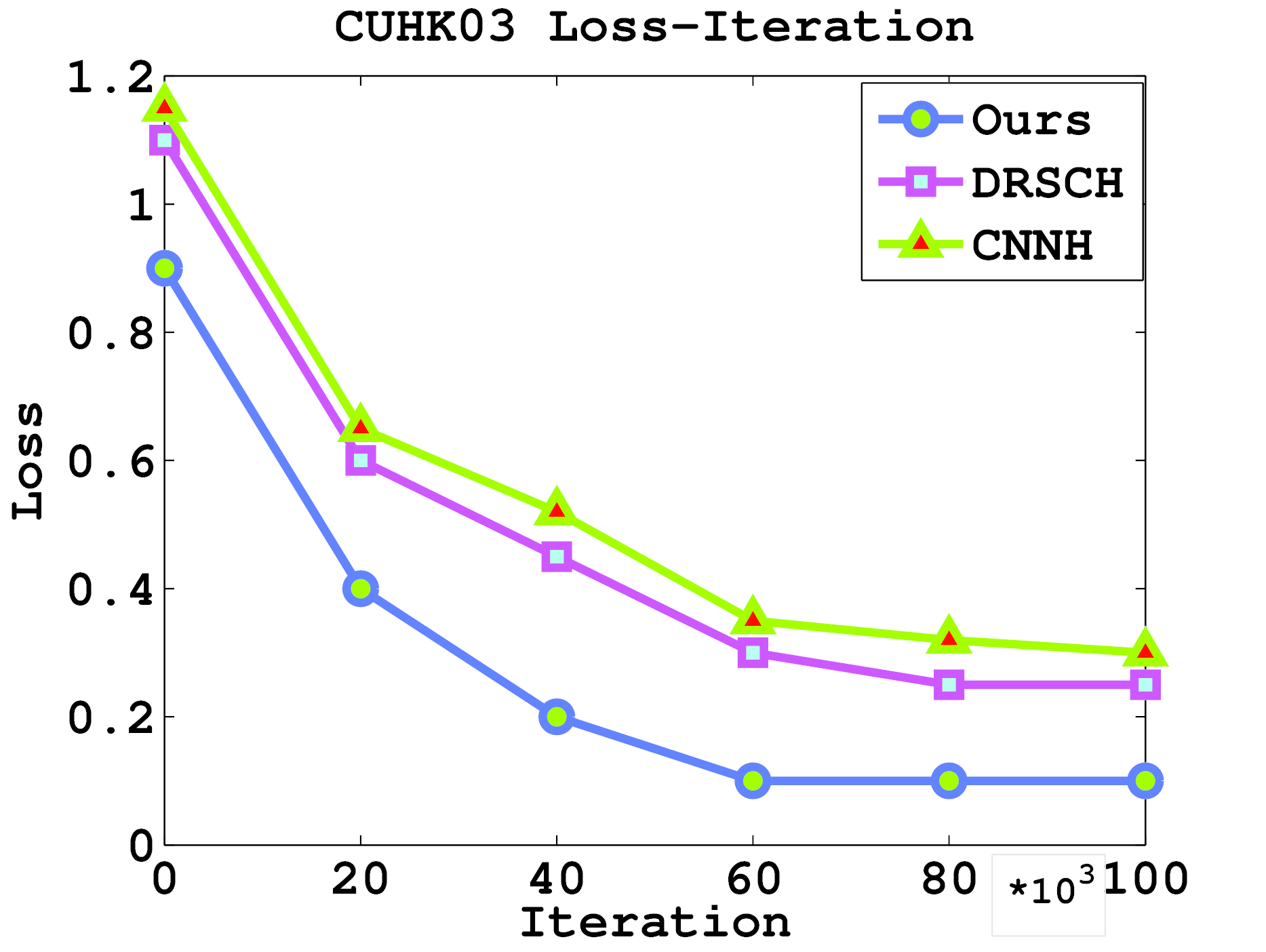}
	\includegraphics[width=1.7in,height=1.3in]{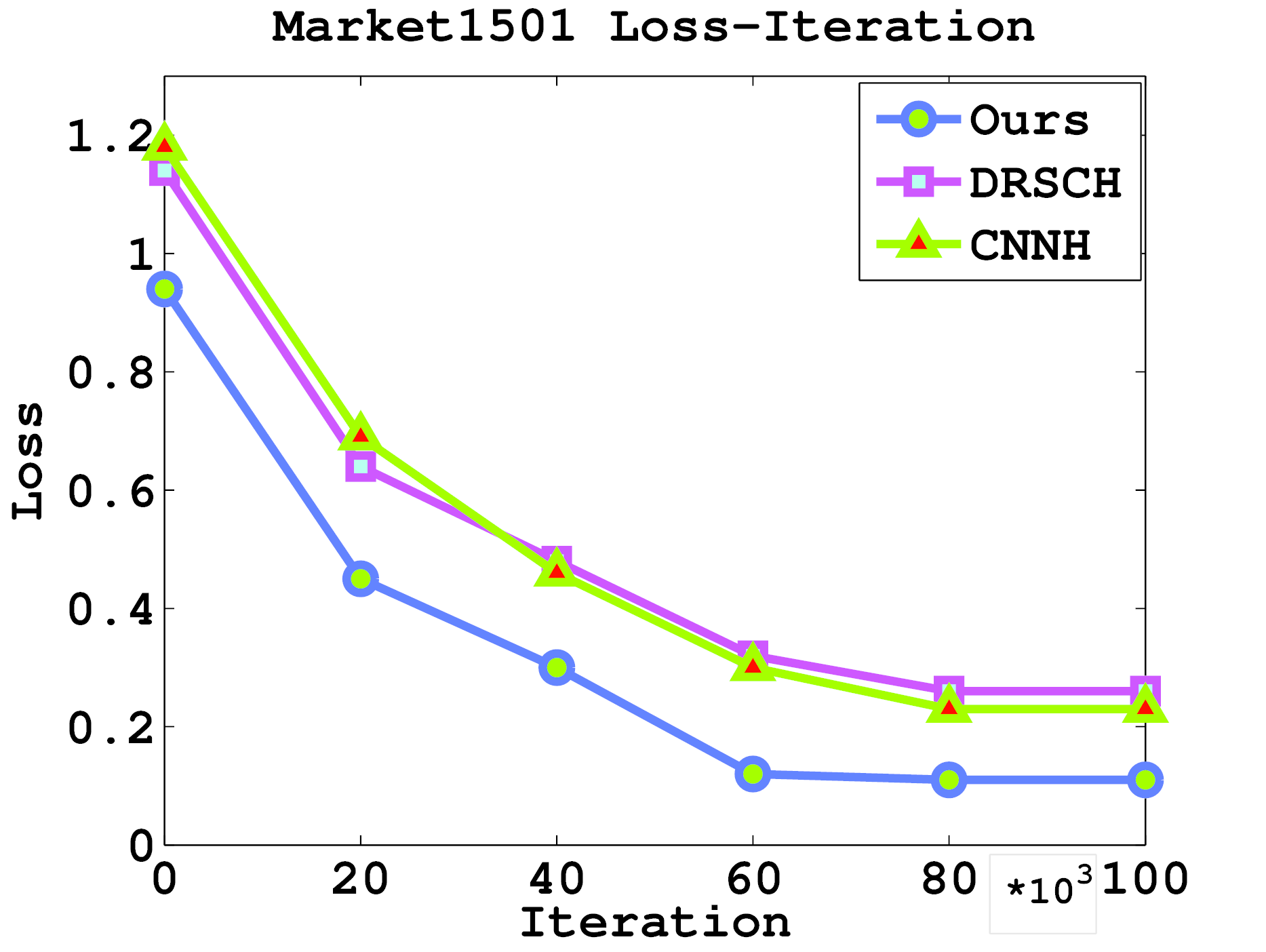}\\
(a) \hspace{4cm} (b)\\
    \caption{Convergence study on two benchmark datasets. It is obvious that our structured embedding has fast convergence compared with contrastive and triplet embeddings.}\label{fig:convergence}
\end{figure}

In this experiment, we study the convergence speed of optimizing contrastive, triplet, and structured embedding, respectively. The average loss values over all mini-batches are computed on three kinds of embeddings, as shown in Fig.\ref{fig:convergence}. We can see that the proposed structured embedding is able to converge faster than other two embeddings. This can be regarded as the response to the augment from hard negatives which provide informative gradient direction for positives.

\section{Conclusion}\label{sec:con}
In this paper, we developed a structured deep hashing architecture for efficient person re-identification, which jointly learn both CNN features and hash functions/codes. As a result, person re-identification can be resolved by efficiently computing and ranking the Hamming distances between images. A structured loss function is proposed to achieve fast convergence and more stable optimization solutions.
Empirical studies on two larger benchmark data sets demonstrate the efficacy of our method. In our future work, we would explore more efficient training strategies to reduce training complexity, and possible solutions include an improved loss function based on local distributions.

\bibliographystyle{named}
\bibliography{ijcai16}

\end{document}